\documentclass[10pt,twocolumn,letterpaper]{article}

\usepackage[accsupp]{axessibility}  
\usepackage{iccv}
\usepackage{times}
\usepackage{epsfig}
\usepackage{graphicx}
\usepackage{amsmath}
\usepackage{amssymb}
\usepackage[normalem]{ulem}
\usepackage{enumerate}
\usepackage{multirow}
\usepackage{verbatim}
\usepackage{booktabs}
\usepackage{pifont}
\usepackage[font=small,labelfont=bf]{caption}
\usepackage[dvipsnames]{xcolor}

\usepackage[numbers,sort&compress]{natbib}
\usepackage{etoolbox}
\makeatletter
\patchcmd{\NAT@citexnum}
  {%
    \ifx\NAT@last@yr\relax
      \def@NAT@last@yr{\@citea}%
    \else
      \def@NAT@last@yr{--\NAT@penalty}%
    \fi
  }
  {%
    \def@NAT@last@yr{--\NAT@penalty}%
  }
  {}{\FAIL}
\makeatother

\usepackage[pagebackref=true,breaklinks=true,letterpaper=true,colorlinks,bookmarks=false]{hyperref}

 \iccvfinalcopy 

\newcommand{\OURS}{PHRIT} 
\newcommand{\cmark}{\textcolor{OliveGreen}{\ding{52}}}%
\newcommand{\xmark}{\textcolor{Maroon}{\ding{56}}}%
\newcommand{\revise}[1]{\textcolor{black}{#1}}

\def\iccvPaperID{2105} 

\ificcvfinal\fi

\let\oldtwocolumn\twocolumn
\renewcommand\twocolumn[1][]{%
   \oldtwocolumn[{#1}{
    \setlength{\abovecaptionskip}{0.cm}
    \vspace{-0.3in}
    \begin{center}
    \centering
    \includegraphics[width=\textwidth]{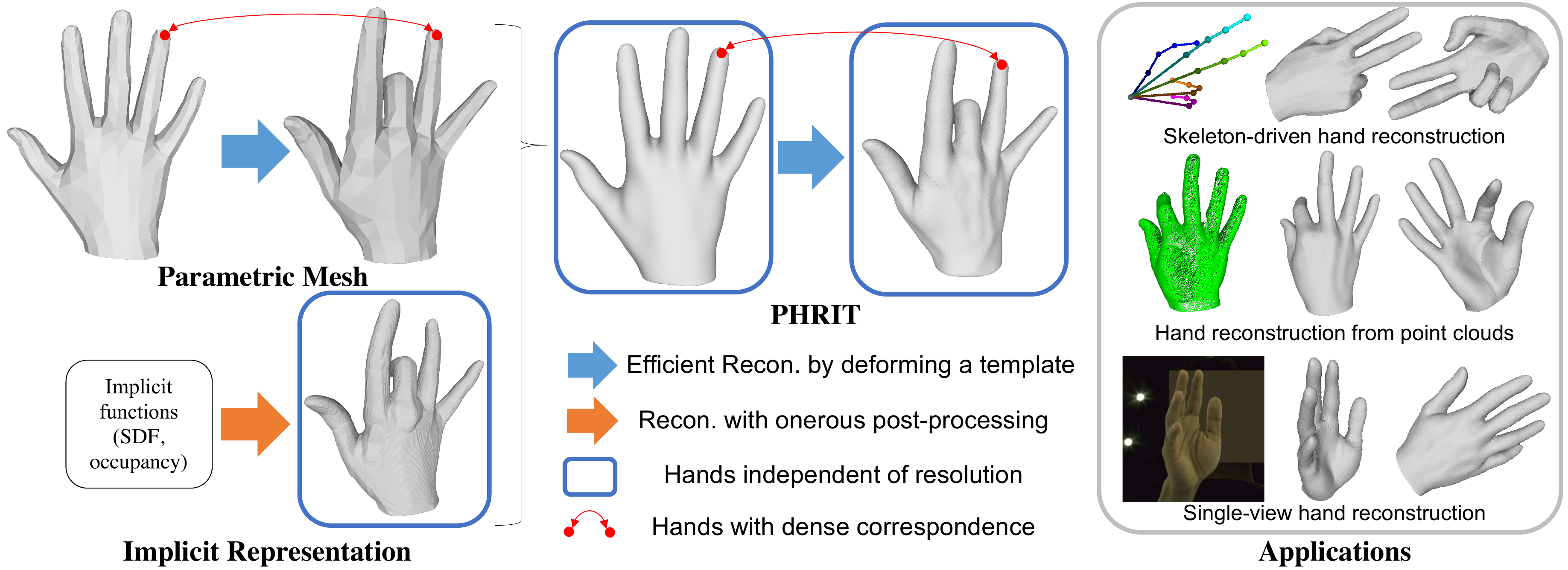}
    \vspace{-0.2em}
    \vspace{-0.4em}
    \captionof{figure}{We propose a novel hand representation \OURS{}, combing advantages of previous methods on hand geometry modeling (namely parametric mesh and implicit representation). \OURS{} can serve as a learned fully differentiable layer driven by input hand skeleton and shape latent code, making it easily applicable to downstream tasks.}
    \label{fig:teaser}
\end{center}
}]
}

\begin{document}

\title{PHRIT: Parametric Hand Representation with Implicit Template}
\author{
Zhisheng Huang\textsuperscript{1}$^{\dagger}$ \quad
Yujin Chen\textsuperscript{2}$^{\dagger}$ \quad
Di Kang\textsuperscript{3} \quad
Jinlu Zhang\textsuperscript{1} \quad
Zhigang Tu\textsuperscript{1}$^*$ \\
\textsuperscript{1}Wuhan University \quad
\textsuperscript{2}Technical University of Munich \quad
\textsuperscript{3}Tencent AI Lab \\
}

\maketitle
\ificcvfinal\thispagestyle{empty}\fi

\begin{abstract}
\vspace{-1em}

We propose \OURS{}, a novel approach for parametric hand mesh modeling with an implicit template that combines the advantages of both parametric meshes and implicit representations.
Our method represents deformable hand shapes using signed distance fields (SDFs) with part-based shape priors, utilizing a deformation field to execute the deformation.
The model offers efficient high-fidelity hand reconstruction by deforming the canonical template at infinite resolution. Additionally, it is fully differentiable and can be easily used in hand modeling since it can be driven by the skeleton and shape latent codes.
We evaluate \OURS{} on multiple downstream tasks, including skeleton-driven hand reconstruction, shapes from point clouds, and single-view 3D reconstruction, demonstrating that our approach achieves realistic and immersive hand modeling with state-of-the-art performance.

\end{abstract}

\newcommand\blfootnote[1]{
\begingroup
\renewcommand\thefootnote{}\footnote{#1}
\addtocounter{footnote}{-1}
\endgroup
}
\blfootnote{$^\dagger$Equal contributions.}  
\blfootnote{$^*$Corresponding author.} 

\vspace{-1.5em}
\section{Introduction}
The human hand plays a vital role in communication and interaction, making high-fidelity hand modeling crucial for immersive applications, especially in the era of digital twins and metaverses. High-fidelity hand modeling can facilitate immersive applications such as virtual meetings and video games. However, to drive these real-time applications, it's essential to achieve realistic reconstruction results, ensure stable cross-user generalization, and optimize the reconstruction process for efficiency.

Previous research on hand geometry modeling can be divided into two broad categories: parametric meshes and implicit representations.
Parametric meshes rely on a predefined mesh template that is deformed to match posed hands \cite{li2022nimble, romero2022embodied, moon2020deephandmesh}. 
While this approach is efficient and provides useful dense correspondence between the reconstructed hand and the canonical template \cite{hasson2021towards,qian2020html}, it requires careful supervision to learn the deformation of each vertex.
This can be difficult and expensive, as noted in  \cite{li2022nimble}.
This challenge has led previous work to either sacrifice resolution \cite{romero2022embodied} or resort to leveraging weak supervisions \cite{moon2020deephandmesh}, which may limit the model's generalization to personalized settings.
In contrast, recent implicit representations \cite{karunratanakul2021skeleton, corona2022lisa} take a different approach to hand geometry modeling by focusing on the continuous representation of static shapes.
By learning implicit functions such as signed distance and occupancy fields, they can represent high-fidelity, resolution-independent shapes.
However, the method requires time-consuming post-processing to obtain reconstructions and lacks dense correspondence between the reconstructions.
Overall, both approaches have their strengths and weaknesses, and neither fully meets the requirements for high fidelity, generalization ability, and efficiency.

\begin{table}[t]
\centering
\scalebox{0.85}{
\begin{tabular}{c|cccc}
\toprule
Method & \textit{Diff.  } & \textit{Effi.  } & \textit{Corres.} & \textit{ Conti.} \\ 
\midrule
Parametric Hand Meshes & \cmark & \cmark & \cmark & \xmark\\ 
Implicit Hand Representation & \cmark & \xmark  &  \xmark & \cmark  \\ 
\OURS{} (Ours)    & \cmark & \cmark & \cmark & \cmark  \\ 
\bottomrule
\end{tabular}
}
\vspace{-0.1in}
\caption{
A comparison of \OURS{} with parametric hand meshes and implicit hand representation in terms of some key properties.
\textit{Diff.}: whether the reconstruction is differentiable.
\textit{Effi.}: whether reconstruction is efficient without onerous pose-processing such as Marching Cubes \cite{lorensen1987marching}. \textit{Corres.}: whether dense correspondences are maintained during reconstruction.
\textit{ Conti.}: whether this representation is continuous in the output space.
}
\vspace{-0.2in}
\label{tab:intro}
\end{table}

To combine the advantages of both paradigms, we propose \OURS{} (as shown in Fig.~\ref{fig:teaser}), a parametric hand model with an implicit template that can generate high-fidelity hand reconstructions for various poses and identities (i.e., generalization ability) with favorable properties such as differentiation, correspondence, efficiency, and continuity.
As summarized in Table~\ref{tab:intro}, our proposed method combines some key strengths of existing paradigms.
We argue that our method achieves continuity to parameterized hand representations by introducing an implicit representation (i.e., using SDF to represent the hand) while maintaining efficiency during inference and maintaining the dense correspondences through our novel deformation field to learn per-vertex deformation implicitly.

Specifically, \OURS{} learns to deform a canonical hand with \revise{theoretically} unlimited resolution based on implicit representation.
To achieve efficiency and continuity (i.e., high fidelity), we represent the canonical hand with an SDF using a neural network.
This means that the canonical hand mesh only needs to be extracted once for all inferences. 
Along with the learning of implicit canonical hand, MLPs are utilized to retrieve per-vertex deformation of the canonical hand conditioned on both poses and shapes.
To learn such deformation,  we use real-world 3D hand scans \cite{romero2022embodied} and develop a novel deformation field that bridges the SDF of deformed and canonical hand space to build dense correspondences (i.e., per-vertex deformation) implicitly. 
To improve the generalization towards unseen poses and identities, we adopt a part-based design \cite{deng2020nasa} on deformation learning, but rather eliminate the requirement of ground-truth bone transformation by deriving local coordinate systems directly upon the hand skeleton based on \cite{karunratanakul2021skeleton}. Additionally, we adopt the locally pose-aware design similar to \cite{chen2021snarf} on deformation learning to boost generalization.
Moreover, we propose a \revise{skip-connection} structure, which is experimentally proven to be more effective in capturing the non-rigidity of the deformation.

In summary, our contributions are:
\begin{itemize}
\vspace{-0.2cm}
\setlength{\itemsep}{0pt}
\setlength{\parsep}{0pt}
\setlength{\parskip}{0pt}
\setlength{\topsep}{0pt}
\setlength{\partopsep}{0pt}
\item We introduce a neural hand model combining parametric meshes and implicit representations to efficiently produce high-fidelity hand reconstruction with full differentiability to the skeleton and shape latent codes.
\item We develop a novel deformation field to learn pointwise deformation and implicitly establish a dense correspondence between the canonical hand and its deformed shape.
\item Experiments demonstrate our method achieves state-of-the-art performance in multiple hand modeling tasks, including reconstruction from skeleton, point clouds, and images, resulting in immersive results.
\end{itemize}

\section{Related Work}

\begin{figure*}[t]
    \vspace{-0.2in}
    \includegraphics[width=\linewidth]{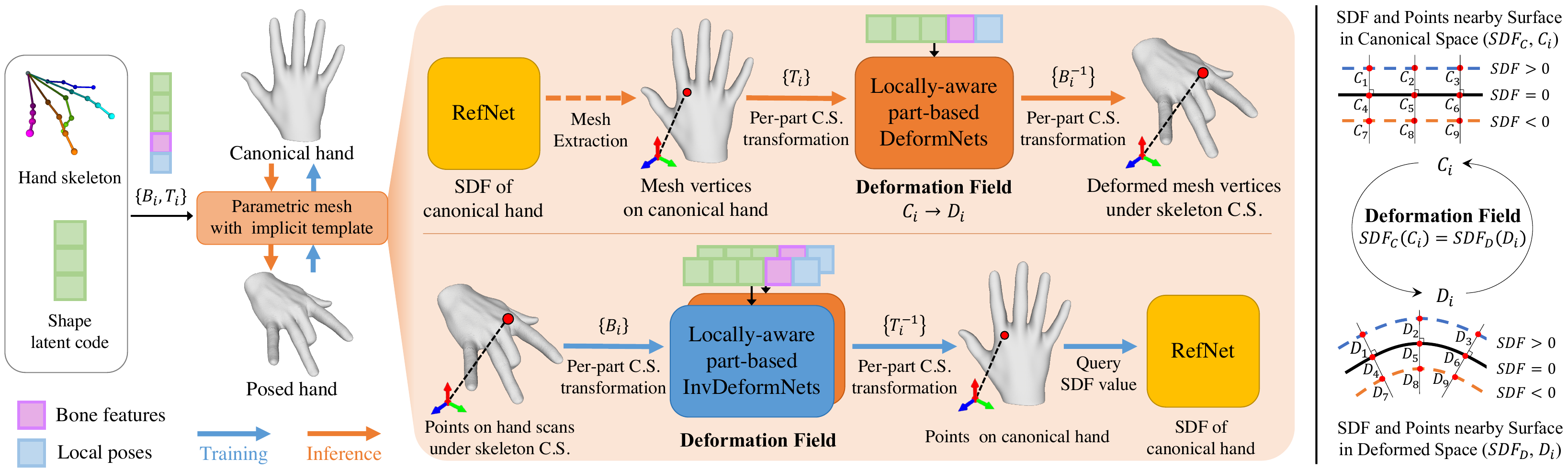}
    \vspace{-0.2in}
   \caption{Overiew of \OURS{} (Section~\ref{sec:pmit}). 
   During training, \OURS{} learns a one-to-one mapping of query points between deformed space and canonical space based on the proposed deformation field (Section~\ref{sec:ldf}), given the hand skeleton and shape latent code. During inference (Section~\ref{sec:isw}), \OURS{} deforms the high-resolution canonical hand mesh extracted from the implicit representation to obtain a differentiable, high-fidelity hand reconstruction. We use a dashed line to denote that mesh extraction is only necessary once.
   }
\label{fig:pipeline}
\vspace{-0.5cm}
\end{figure*}

\noindent \textbf{Parametric hand mesh.} 
Parametric meshes are widely used to model deformable shapes such as faces \cite{blanz1999morphable, li2017learning}, bodies \cite{loper2015smpl, osman2020star, pavlakos2019expressive, joo2018total}, hands \cite{moon2020deephandmesh, li2022nimble, chen2021model, tu2023consistent}, feet \cite{luximon2004foot}, and animals \cite{zuffi20173d}.
While MANO \cite{romero2022embodied} is the most commonly used hand model \cite{boukhayma20193d, park2022handoccnet, chen2021self, chen2021model, chen2021joint, yang2021cpf, yang2022oakink, hampali2020honnotate, moon2020interhand2, hasson2019learning, taheri2020grab, corona2020ganhand}, it has limited resolution and a linear representation of non-rigid deformation, leading to unrealistic hand reconstructions.
To overcome these limitations, previous works have explored graph networks \cite{choi2020pose2mesh, ge20193d, tang2021towards, li2022interacting}, transformers \cite{lin2021end, cho2022cross}, and UV maps \cite{chen2021i2uv}.
DHM \cite{moon2020deephandmesh} proposes to model high-fidelity hand meshes by deforming a professionally designed template and predicting pose- and shape-dependent correctives through neural networks, but with a personalized setup.
Inspired by MANO and DHM, our generalizable canonical hand template mesh addresses resolution limitations by using an implicit representation (i.e.,~SDF) and allows for non-rigid deformation of vertices.

\noindent \textbf{Implicit hand representation.} 
Implicit representations, like SDF \cite{park2019deepsdf}, occupancy \cite{mescheder2019occupancy}, and neural radiance fields \cite{mildenhall2021nerf}, are widely used in 3D modeling  because they are continuous and differentiable \cite{chen2019learning, hong2022avatarclip, karunratanakul2020grasping, karunratanakul2021skeleton, peng2020convolutional, niemeyer2020differentiable, saito2019pifu, ma2022neural, mildenhall2021nerf, oechsle2020learning, peng2021neural, zhang2020nerf++, mihajlovic2021leap, tang2021sa, tang2021skeletonnet}.
Recently, implicit representations have been applied to articulated shapes in \cite{bozic2020deepdeform, feng2022capturing,peng2021animatable,niemeyer2019occupancy,noguchi2021neural,palafox2022spams, lombardi2021latenthuman, alldieck2021imghum, tang2022neural}. 
These works use implicit skinning fields \cite{saito2021scanimate} or piecewise deformable models \cite{deng2020nasa} to represent articulated objects.
Our work draws inspiration from these previous works and proposes a novel solution for modeling high-fidelity articulated hands. 
\revise{In contrast to Mehta \textit{et al.} \cite{mehta2022level}, who maintain differentiability with Marching Cubes to enable supervision for explicit representation and update the geometry of the implicit surface, we directly impose supervision on the implicit representation.}
Unlike previous methods using implicit skinning functions to blend a fixed shape \cite{saito2021scanimate,chen2021snarf,jeruzalski2020nilbs}, our model is conditioned on pose and shape variation. 
Compared to methods \cite{palafox2021npms,palafox2022spams,yenamandra2021i3dmm} using implicit functions to encode shape variation and pose-dependent deformation separately, our model is more efficient and shares correspondences across identities.
Furthermore, we address the multiple correspondence problem in SNARF \cite{chen2021snarf} by developing a deformation field and introducing a deformation \revise{skip-connection} structure to learn it.

\noindent \textbf{Skeleton-driven articulated hand.}  
Although previous studies \cite{choi2020pose2mesh,ge20193d} have demonstrated the feasibility of skeleton-driven shape reconstruction, recent research \mbox{\cite{corona2022lisa,deng2020nasa,noguchi2021neural}} on articulated shape modeling still relies on ground truth bone transformations. This is mainly because bone transformations provided by iterative optimization procedures like inverse kinematics prevent gradient flow and introduce ambiguity in the twist angle of bones. 
To address this issue, \cite{karunratanakul2021skeleton} exploits biomechanical constraints \cite{spurr2020weakly} to define the local coordinate systems of hand skeletons and derive their bone transformation accordingly. 
However, their proposed approach involves learning deformations in canonical space and requires additional canonicalization of the hand skeleton.
In contrast, we directly learn the deformations in each hand's local coordinate system via more readily available transformation matrices.
\section{Approach}

We present a high-fidelity hand model \OURS{} (Section~\ref{sec:pmit}). 
As shown in Fig.~\ref{fig:pipeline}, \OURS{} consists of RefNet and DeformNets, which are jointly learned through the proposed deformation field (Section~\ref{sec:ldf}) with an inverse counterpart of DeformNets, denoted as InvDeformNets. The resulting \OURS{} allows for efficient and high-fidelity hand reconstruction that is fully differentiable with respect to both the input hand skeleton and shape latent code (as explained in Section~\ref{sec:isw}). 

\subsection{Parametric Mesh with Implicit Template}
\label{sec:pmit}
Our \OURS{} comprises an implicit template (RefNet) and an associated deformation function (DeformNets).

\vspace{-1em}
\subsubsection{RefNet}
\label{sec:refnet}
\vspace{-0.3em}
RefNet $f$ encodes a canonical hand mesh $M$, which serves as the basis for all hand reconstructions generated by our model. 
We define $M$ to be compatible with the MANO \cite{romero2022embodied} template mesh $\bar{M}$ (of zero pose and mean shape). That is, $M$ and $\bar{M}$ have the same skeleton $\bar{K}$ and shape, but $M$ has an infinite number of mesh vertices. RefNet $f$ maps a query point x to its signed distance $d$ to the hand surface: $f(x) = d \in \mathbb{R}$.
 
\vspace{-1.22em}
\subsubsection{DeformNets}
\vspace{-0.4em}
Our deformation function utilizes a part-based design, where we decompose the human hand into 16 rigid parts $\left \{ P_i \mid 0\le i \le 15 \right \} $ based on the MANO \cite{romero2022embodied} skinning weights. 
For each part $P_i$, we employ an independent DeformNet $g_i$ conditioned on a local pose $\theta_i$ and a shape code $\beta_i$, transforming the query points $x_i$ on $P_i$ of canonical hand to corresponding points $y_i$ on $P_i$ of deformed hand: $g_{i}\left ( x_i, \theta_i, \beta_i  \right ) = y_i \in \mathbb{R}^{3}$.

To define the query points $x_i$ in a local coordinate system $LCS_i$ of the part $P_i$, we follow NASA's approach \cite{deng2020nasa}. However, we derive the $LCS_i$ from the hand skeleton rather than using ground truth bone transformations.

\begin{figure}[t]
\vspace{-0.5cm}
\includegraphics[width=\linewidth]{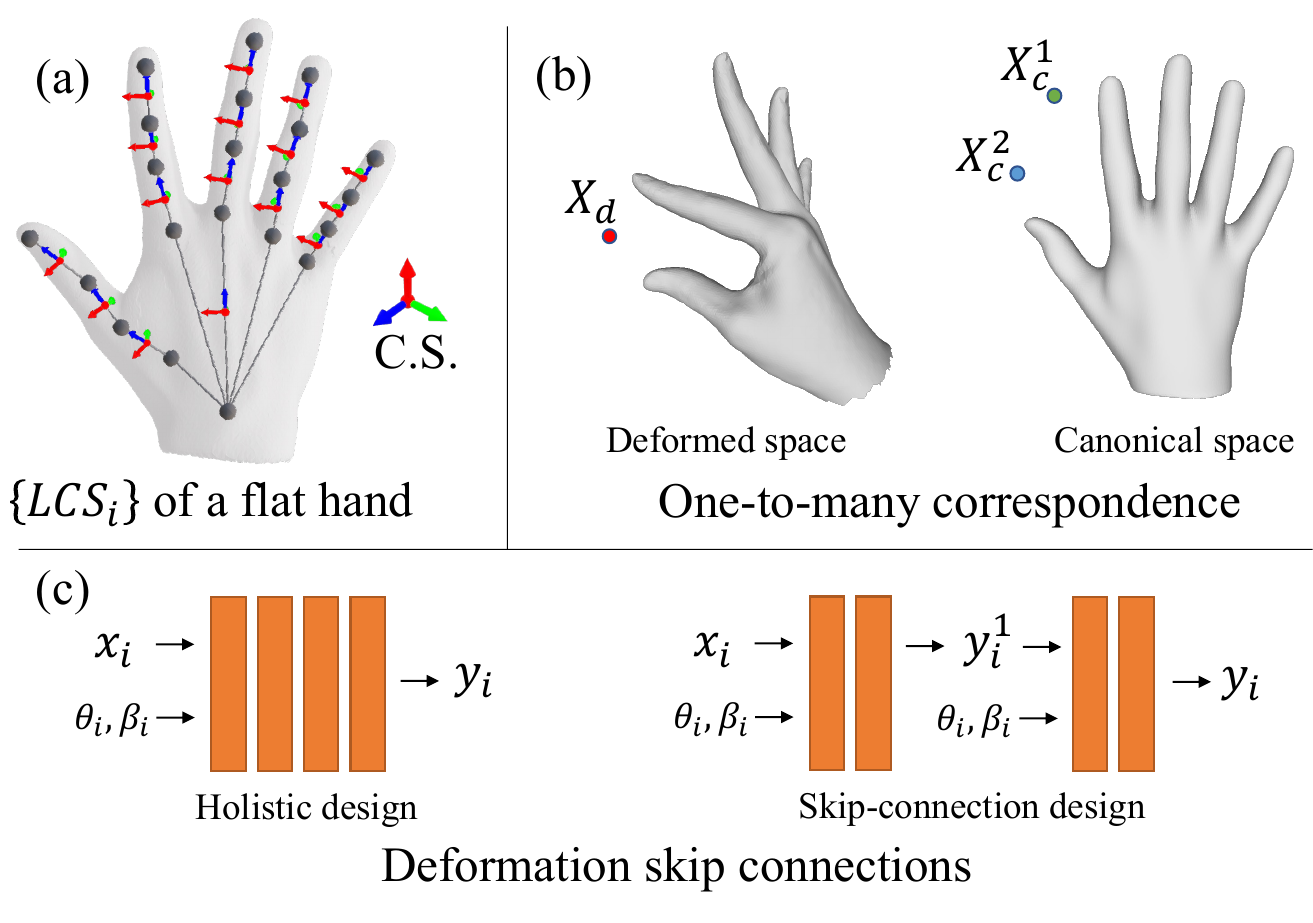}
\vspace{-0.3in}
\caption{(a) Local coordinate systems derived from hand skeleton. (b) Multi-correspondence (one-to-many) between $X_d$ and $X_c$ (Section~\ref{sec:ldf}). (c) Deformation \revise{skip-connection} architecture with comparison to the holistic design.}
\label{fig:misc}
\vspace{-0.5cm}
\end{figure}

\noindent \textbf{Local coordinate system $LCS_i$.} Given a hand skeleton consisting of 21 3D keypoints and 20 bones $\left \{ b_j \mid 1\le j \le 20 \right \} $, we first define a transformation $B_j$ for each bone $b_j$ and then define the $LCS_i$ for each of the 16 rigid hand parts $P_i$. 
For $B_j$ of bone $b_j$, the translation is set to the middle point of $b_j$, and the rotation (orientation) is set following HALO \cite{karunratanakul2021skeleton}.
Details can be found in the Appendix.
Based on $B_j$, we construct the local coordinate system $LCS_i$ for each part $P_i$ as follows: (1) For palm part $P_1$, $LCS_1$ is defined by $B_9$ (i.e., the transformation of the middle palmar bone), where the origin is set to the translation of $B_9$ and the axes are set to the orientation of $B_9$. (2) For the remaining finger parts $P_i$, $LCS_i$ is defined by corresponding finger bone transformation $B_j$, in the same way as $LCS_1$ is determined by $B_9$. An illustration of $LCS_i$ is provided in Fig.~\ref{fig:misc} (a).

\noindent \textbf{Local pose $\mathbf{\theta_i}$.} The local pose $\theta_i$ of $P_i$ is defined differently for the palm part $P_1$ and the remaining finger parts $P_i$.
(1) For palm part $P_1$, the $\theta_1 \in\mathbb{R}^{17}$ consists of a palmar bone configuration $\theta_1^p \in \mathbb{R}^{7}$ and palmar joint configuration $\theta_1^f \in \mathbb{R}^{10}$. 
Following HALO \cite{karunratanakul2021skeleton}, $\theta_1^p$ is decoupled into a finger spreading $\theta_s \in \mathbb{R}^4$ and a palm arching $\theta_a \in \mathbb{R}^3$.
$\theta_s$ is defined by angles between adjacent palmar bones, and $\theta_a$ is defined by angles between the two planes spanned by three adjacent palmar bones. The $\theta_1^f$ are joint angles (abduction and flexion) of each five palmar joints. 
(2) For the remaining finger parts $P_i$, $\theta_i$ are joint angles from the corresponding finger bones. 
 Each finger bone has two joints with two joint angles, namely abduction and flexion, except for the tips of level-2 finger bones, which have no joint angles.
 
\noindent \textbf{Shape code $\beta_i$.} Similar to SNARF \cite{chen2021snarf}, we split shape space into surface properties and bone features. \revise{Formally, the shape code $\beta_i = \gamma \oplus F_i$, where $i$ indexes hand parts and $\oplus$ denotes feature concatenation.  $\gamma \in \mathbb{R}^{128}$ is a trainable shape latent code shared across all hand parts, which is optimized per hand ID (subscript omitted), and $F_i$ is bone features from the hand part $P_i$.}
Following HALO \cite{karunratanakul2021skeleton}, $F_i$ is a concatenation of global bone features $F_g \in \mathbb{R}^{16}$ obtained from a global bone encoder and local bone lengths $L_i$ of $P_{i}$.
Except for $L_1 \in \mathbb{R}^{5}$, which is set to the lengths of five palmar bones, the remaining $L_i \in \mathbb{R}$ is the length of the corresponding finger bone. 

\noindent \textbf{Deformation \revise{skip connections}.}
We propose a \revise{skip connection} design for learning our DeformNets $g_i$ instead of the holistic design depicted in Fig.\ref{fig:misc}.
The $g_i$ is decomposed as $g_{i} = g_i^0 \circ g_i^1 \cdots  \circ g_i^N$,  where $g_i^{n+1}$ takes the deformation results $y_i^n$ from $g_i^{n}$ to predict further deformation: $g_{i}^{n+1}\left ( y_i^{n}, \theta_i, \beta_i  \right ) = y_i^{n+1} \in \mathbb{R}^{3}$. $N$ is the number of  \revise{skip connections} and $\theta_i$ and $\beta_i$ are the local pose and shape code as before.
We empirically find that the  \revise{skip-connection} design, with the same total number of layers, better captures the details in hand reconstruction. (See Section \ref{sec:ablation} for an ablation study on this design choice.)

\subsection{Learning with Deformation Field}
\label{sec:ldf}
To learn the RefNet and DeformNets introduced in the previous section, we propose a novel deformation field and derive training objectives based on it.

\noindent \textbf{Deformation field $\phi$.} The proposed deformation field is a one-to-one mapping function that transforms a 3D point $X_c$ in the canonical hand space to $X_d$ in deformed hand space: $\phi: X_c \mapsto X_d$. 
To exclude potential ambiguity arising from multi-correspondence between $X_d$ and $X_c$, as pointed by SNARF \cite{chen2021snarf} (Refer to Fig.~\ref{fig:misc} (b)), we define $\phi$ as folows: (1) $\phi$ has a constrained domain nearby the hand surface. (2) For $X_c$ on the hand surface, $\phi$ is based on the correspondence between canonical hand surface and the deformed hand surface. (3) For $X_d$ off the hand surface but within the domain, $\phi$ is further based on the signed distance to the hand surface. A 2D illustration is provided on the right side of Fig.~\ref{fig:pipeline}. The resulting $\phi$ satisfies the following equation:
\begin{equation}
\label{eq:df}
\begin{split}
    SDF_d&(X_d) =SDF_d(\phi(X_c)) =SDF_c(\phi^{-1}(X_d))
\end{split}
\end{equation}
Here, $SDF_c$ and $SDF_d$ are signed distance fields of the hand in canonical space and deformed space, respectively. A more concrete mathematical definition of $\phi$ and its constrained domain can be found in Appendix.

\noindent \textbf{Training objectives.} Our training objectives are based on the deformation field $\phi$. 
Specifically, our proposed DeformNets $g_i$ follow $\phi$ as an invertible mapping function, and we introduce an inverse counterpart of DeformNets called InvDeformNets $g_{i}^{-1}$.
The $g_i$ and $g_{i}^{-1}$, along with the RefNet $f$, are jointly learned using the following loss function:
\begin{equation}
\begin{split}
    E = &\sum_{s=1}^B \sum_{i=1}^{P} \sum_{x_i^s \in X_i^s} (w_S E_{SDF} + w_n E_{norm} + w_c E_{cycle} \\
    & + w_m E_{mano} + w_r E_{regu}) + w_O E_{O} + w_I E_{IGR} 
\end{split}
\end{equation}
Here, $B$ is the number of scans in the batch, $P=16$ is the number of hand parts, and $\left \{ X_i^s \mid x_i^s \in \mathbb{R}^3  \right \}$ are hand surface points on the part $P_i$ in the $s$-th scan.
The individual loss terms are defined as follows:
\begin{equation}
   E_{SDF} = \left \| f( T_i^{-1} \cdot  g_{i}^{-1}(x_i^s, \theta_i^s, \beta_i^s)) \right \|
\end{equation}
\begin{equation}
     E_{norm} = \left \|\nabla_{x_i^s}  f( T_i^{-1} \cdot  g_{i}^{-1}(x_i^s, \theta_i^s, \beta_i^s))- N(x_i^s) \right \| 
\end{equation}
\begin{equation}
    E_{cycle} = \left \| g_{i}(g_{i}^{-1}(x_i^s, \theta_i^s, \beta_i^s),\theta_i^s, \beta_i^s)- x_i^s \right \| 
\end{equation}
\begin{equation}
    E_{mano} = \left \| g_{i}^{-1}(\bar{x}_i^s, \theta_i^s, \beta_i^s) - \bar{y}_i^s \right \|  
\end{equation}
\begin{equation}
    E_{regu} = \left \| \gamma^s \right \| 
\end{equation}
\begin{equation}
    E_{O} = \mathbb{E}_{x\in \Omega\setminus\Omega_0 } \space (\exp(-\alpha \cdot \left | f(x)\ \right | ) ), \quad  \alpha \gg 1
\end{equation}
\begin{equation}
   E_{IGR} = \mathbb{E}_{x\in \Omega} (\left \| \nabla f(x) \right \| - 1)^{2}
\end{equation}

$E_{SDF}$ ensures the points $x_i^s$ on deformed hand surface have zero signed distance. \revise{We query the SDF of $x_i^s$ through the deformation field $\phi$. Specifically}, we first deform $x_i^s$ using $g_{i}^{-1}$, then transform it back from the $LCS_i$ of canonical hand using $T_{i}^{-1}$, and finally query the signed distance using $f$.

\revise{$E_{norm}$ further constrains the normal of the points $x_i^s$ on deformed hand surface based on $\phi$, which satisfies Eq.\ref{eq:df}.}
This constraint is based on the following two observations: (1) The derivative of SDF is the corresponding gradient field, and (2) The SDF of deformed hand space is associated with the SDF of canonical hand space by Eq.\ref{eq:df}. 
We denote the surface normal of $x_i^s$ with $N(x_i^s)$. 

$E_{cycle}$ facilitates that $g_i$ and $g_{i}^{-1}$ are reciprocal functions, \revise{ensuring a one-to-one mapping between the canonical hand and deformed hand surfaces.}

$E_{mano}$ enforces $f$ to learn a canonical hand compatible with the MANO template. To achieve this, we use MANO annotations to provide sparse correspondence supervision. We denote MANO vertices of the canonical hand and deformed hand under $LCS_i$ as $\bar{x}_i^s$ and $\bar{y}_i^s$ respectively.

$E_{regu}$ regularizes the trainable shape latent code $\gamma^s$.

$E_O$ and $E_{IGR}$ are utilized to regularize $f$ to learn a standard SDF. 
$E_{IGR}$ is the Eikonal regularization term proposed  in \cite{gropp2020implicit}, while $E_{O}$ ensures off-surface points do not have zero signed distances, as previously introduced in \cite{sitzmann2020implicit}.  

\subsection{Inference with Skinning Weights}
\label{sec:isw}
During inference, our fully differentiable parametric hand mesh model is driven by an input skeleton $K$ and a shape latent code $\gamma$. The inference pipeline consists of the following steps:
\begin{enumerate}[(1)]
\vspace{-0.25cm}
\setlength{\itemsep}{0pt}
\setlength{\parsep}{0pt}
\setlength{\parskip}{0pt}
\setlength{\topsep}{0pt}
\setlength{\partopsep}{0pt}
    \item  We extract the canonical hand mesh $M$ from RefNet $f$ using Marching Cubes and assign skinning weights to its vertices by upsampling the skinning weights of the MANO template to ensure compatibility with $M$.
    \item  Using $K$ and $\gamma$, we compute $LCS_i$, $\theta_i$, and $\beta_i$. 
    \item DeformNets deforms the vertices of $M$ based on $\theta_i$ and $\beta_i$, and then transforms the deformed vertices to the coordinate system where the skeleton $K$ is located using $LCS_i$. Finally, we combine the transformed vertices based on their skinning weights to obtain the final reconstruction result.
\vspace{-0.25cm}
\end{enumerate}
As our canonical hand mesh represented by RefNet $f$ is static, step (1) is only needed once for all inferences.

\section{Experiments}
\subsection{Datasets and Evaluation Metrics}

\noindent \textbf{Youtube3D (YT3D) \cite{kulon2020weakly}:} 
The YT3D dataset has 50,175 hand meshes from 109 videos, with 47,125 meshes from 102 videos in the training set and 1,525 meshes from 7 videos in the test set. As the YT3D annotations are fitted from video data, we use its MANO pose and mesh for skeleton-driven hand shape reconstruction.

\noindent \textbf{MANO dataset \cite{romero2022embodied}:} The training set contains 1,554 scans with MANO annotations of 31 subjects performing 51 poses, and the test set contains 50 scans of unseen subjects performing unseen poses to the training set. We use the real scans from the MANO trainset to learn our high-fidelity hand mesh model and use the MANO test set to evaluate point cloud 3D hand reconstruction.

\noindent \textbf{DeepHandMesh (DHM) dataset \cite{moon2020deephandmesh}:} The DHM dataset contains multi-view observations including RGB images and depth maps. 3D shape reconstructions and 3D hand keypoints are also provided. 
Since only one subject data is publically accessible, we use this available subset in our experiments, where the released data is split into a training set with 4,550 samples and a test set with 915 samples. We use DHM to learn hand reconstruction from images.

\noindent \textbf{FreiHAND \cite{zimmermann2019freihand}:}
FreiHAND has 130,240 training and 3,960 evaluation samples, all with MANO annotations. Although the MANO annotations are not as accurate as real scans, the dataset is used to showcase how our model can learn high-fidelity single-view hand reconstruction using common yet coarse MANO-level annotations.

\noindent \textbf{Evaluation metrics.}
For skeleton-driven hand reconstruction, we report the mean Intersection over Union (\textbf{IoU}),  Chamfer-L1 distance in $mm$ (\textbf{Cham.}) and normal consistency score (\textbf{Norm.}) in \cite{mescheder2019occupancy}. 
For hand reconstruction from point clouds, we compute vertex-to-vertex (\textbf{V2V} in $mm$) and vertex-to-surface (\textbf{V2S}) distances between the reconstruction and scan in both directions. 
For hand reconstruction from images, the evaluation metrics include 3D joint distance error ($\rm \mathbf{P_{err}}$ in $mm$), mesh vertex error ($\rm \mathbf{M_{err}}$), mean per-joint position error and per-vertex position error with Procrustes analysis \cite{gower1975generalized} (\textbf{PA MPJPE} and \textbf{PA MPVPE} in $mm$), and $F$-score evaluated at thresholds of 5mm and 15mm (\textbf{F@5 mm} and \textbf{F@15 mm}).

\begin{figure}[th]
    \centering
    \vspace{-0.2cm}
    \includegraphics[width=0.9\linewidth]{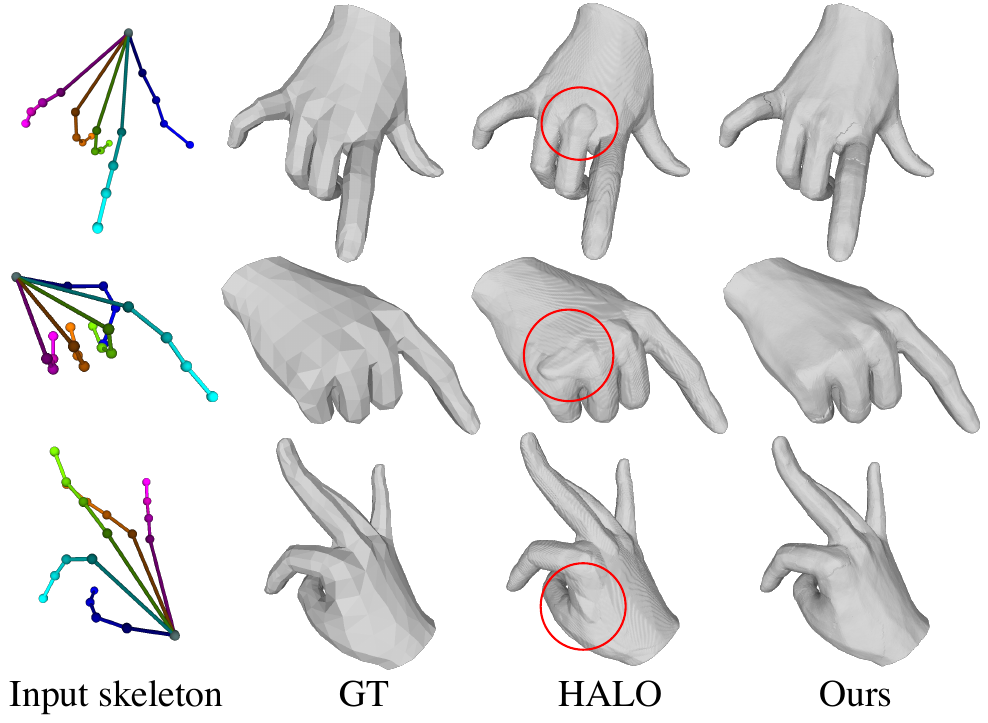}
    \vspace{-0.3cm}
   \caption{Skeleton-driven hand reconstruction results on YT3D. Both HALO and ours maintain the keypoint locations faithfully, while ours achieve more accurate and smoother reconstructions.}
\label{fig:YT3D}
\vspace{-0.5cm}
\end{figure}

\begin{table}[th]
\vspace{-0.2cm}
\centering
\scalebox{1}{
\begin{tabular}{c|ccc}
\hline
Method           & IoU$\uparrow$  & Cham.$\downarrow$ & Norm.$\uparrow$          \\ \hline
NASA \cite{deng2020nasa}          & 0.896          & 1.057          & 0.955          \\
HALO  \cite{karunratanakul2021skeleton}           & 0.932          & 0.719          & 0.959          \\ \hline
HALO(Keypoints) & 0.930          & 0.740          & 0.959          \\
Ours   & \textbf{0.949} & \textbf{0.432} & \textbf{0.979} \\ \hline
\end{tabular}
}
\vspace{-0.3cm}
\caption{Results of skeleton-driven hand reconstruction on YT3D. NASA \cite{deng2020nasa} and HALO \cite{karunratanakul2021skeleton} use GT bone transformation, while HALO(Keypoints) and ours take keypoints as input only.
}
\label{tab:K2M-YT3D}
\vspace{-0.5cm}
\end{table}

\begin{table*}[t]
\centering
\vspace{-0.1in}
\makebox[0pt][c]{\parbox{1\textwidth}{
\begin{minipage}[h]{0.5\textwidth}
    \centering
    \setlength{\abovecaptionskip}{0.cm}
    \centering
    \scalebox{1.1}{
    \begin{tabular}{c|cc|cc}
        \hline
        \multirow{2}{*}{Method}          & \multicolumn{2}{c|}{Recon. to scan} & \multicolumn{2}{c}{Scan to recon.} \\ \cline{2-5}
                    & V2V$\downarrow$             & V2S$\downarrow$            & V2V $\downarrow$             & V2S$\downarrow$        \\ \hline
        MANO \cite{romero2022embodied}      & 3.14                 & 2.92                & 3.90                 & 1.57                \\
        \hline
        VolSDF \cite{yariv2021volume}    & 3.69                 & 2.22                & 2.37                 & 2.23                \\
        NASA \cite{deng2020nasa}     & 5.31                 & 3.80                & 2.57                 & 2.33                \\
        NARF \cite{noguchi2021neural}     & 4.02                 & 2.69                & 2.11                 & 2.06                \\
        LISA-im \cite{corona2022lisa}  & 3.09                 & 1.96                & 1.19                 & 1.13                \\ \hline
        LISA-geom \cite{corona2022lisa} & \textbf{0.36}        & \textbf{0.16}       & 0.81                 & \textbf{0.26}       \\
        LISA-full \cite{corona2022lisa} & 1.45 & 0.64 & \underline{0.64} & 0.58 \\
        Ours      & \underline{0.37}                 & \underline{0.27}                & \textbf{0.37}        & \underline{0.31}         \\ \hline
        \end{tabular}
    }
    \vspace{0.1cm}
    \caption{Results of shape reconstruction from point clouds on the MANO test set. We achieve comparable results to LISA using much fewer training scans.
    }
    \label{tab:PF-MANO}
\end{minipage}
\hfil
\begin{minipage}[h]{0.45\textwidth}
    \centering
    \vspace{-0.3cm}
    \includegraphics[width=0.89\linewidth]{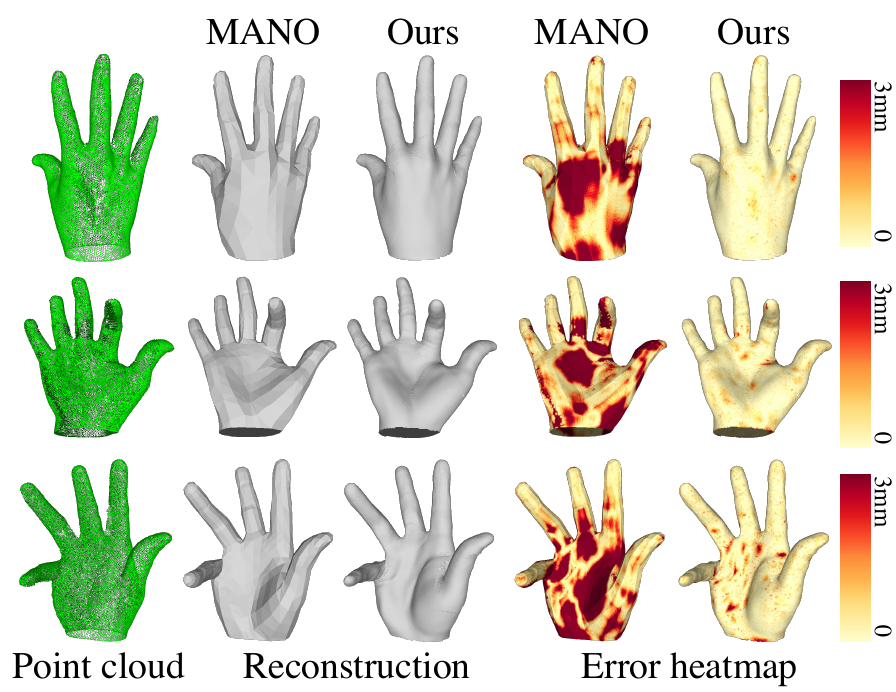}
    \vspace{-0.3cm}
   \captionof{figure}{Results of hand reconstruction from point clouds. Comparisons between MANO and ours are shown.
   }
   \label{fig:MANO_test}
\end{minipage}
}}
\vspace{-0.3cm}
\end{table*}

\subsection{Implementation Details}

Our model is implemented within Pytorch \cite{paszke2019pytorch}, where the Adam solver is used with mini-batches of size 32. The initial learning rate is set to 0.0005, and decayed by a factor of 0.5 after 250 epochs and 500 epochs. We train for 1000 epochs on 2 NVIDIA RTX 2080 Ti GPUs, which takes around 16 hours for training on the MANO dataset. For weighting factors, we set $w_S = 0.1$,  $w_n = 1$,  $w_c = 0.1$, $w_m = 0.1$,  $w_r = 0.0001$,  $w_O = 0.1$ and  $w_I = 1$. 
We refer to the appendix for more details.

\begin{figure*}
\centering
\vspace{-0.4cm}
\includegraphics[width=0.95\linewidth]{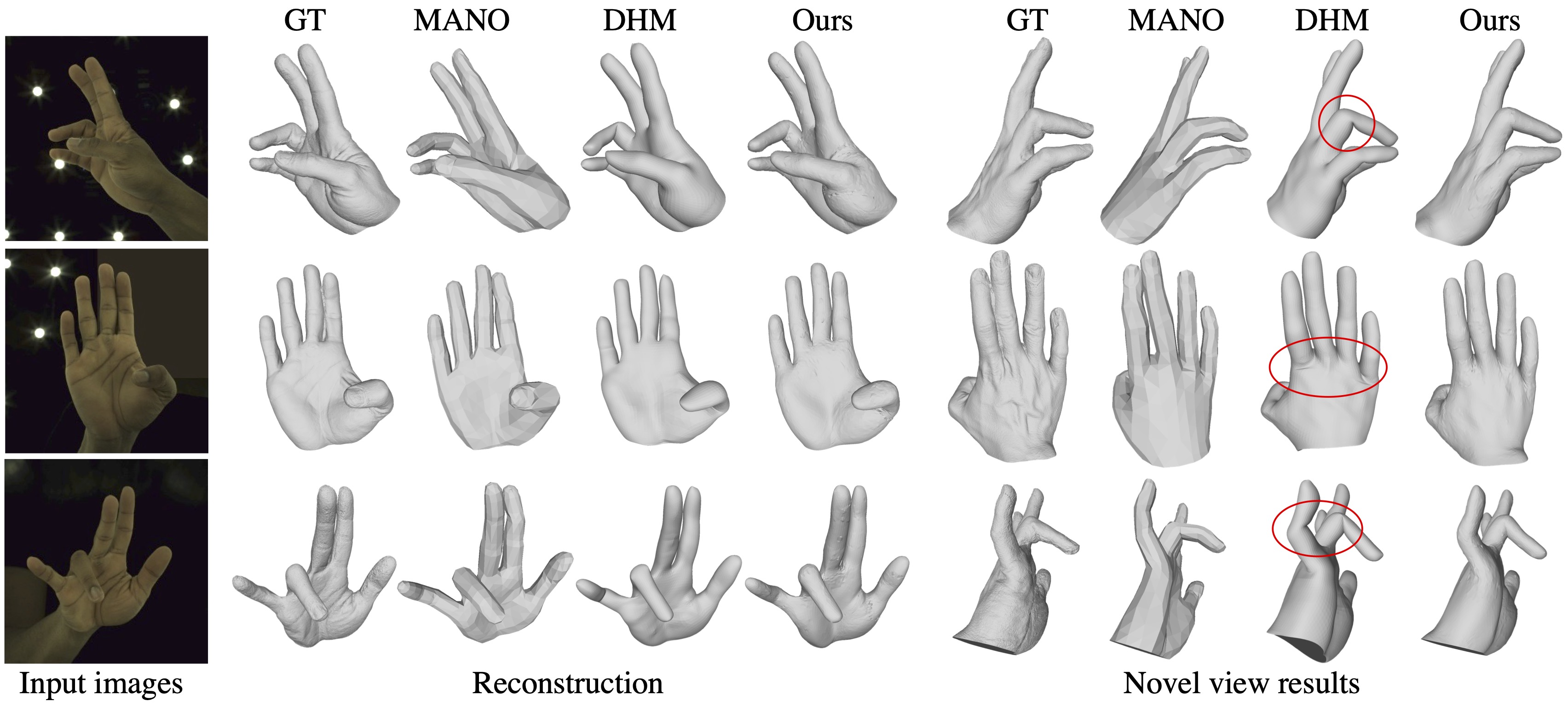}
\vspace{-0.4cm}
\caption{3D reconstruction results from images on DHM. Both DHM \cite{moon2020deephandmesh} and ours reconstruct high-fidelity hands, while ours achieve more natural and accurate results.}
\label{fig:DHM}
\vspace{-0.3cm}
\end{figure*}

\begin{table*}[tb]
\centering
\makebox[0pt][c]{\parbox{1\textwidth}{
\begin{minipage}[h]{0.54\textwidth}
    \centering
    \setlength{\abovecaptionskip}{0.cm}
    \centering
    \scalebox{0.86}{
    \begin{tabular}{c|cccc}
    \hline
    Method  & MPJPE$\downarrow$ & MPVPE$\downarrow$ & F@5 mm$\uparrow$ & F@15 mm$\uparrow$ \\ \hline
    \cite{hasson2019learning} & - & 13.2 & 0.426 & 0.908 \\
    \cite{boukhayma20193d} & - & 13.0 & 0.435 & 0.935 \\
    \cite{zimmermann2019freihand} & - & 10.7 & 0.529 & 0.935 \\
    I2L-MeshNet \cite{moon2020i2l} & \textbf{7.4}   & \textbf{7.6}  & \textbf{0.681}  & \underline{0.973} \\
    \hline
    MANO$^*$ \cite{romero2022embodied}    & 10.8  & 12.4  & 0.485  & 0.947   \\ 
    NIMBLE$^*$ \cite{li2022nimble}   & -  & 9.4  & 0.547  & 0.955   \\ 
    Ours-low$^*$  & \underline{7.5}   & \underline{7.8}   & \underline{0.660}  & \textbf{0.974}   \\
    Ours-high$^*$ & 7.7   & 8.0   & 0.654  & \textbf{0.974}   \\ \hline
    \end{tabular}
    }
    \vspace{0.2cm}
    \caption{Results of shape reconstruction from images on FreiHAND. 
    $^*$To~be consistent with the evaluation in NIMBLE \cite{li2022nimble}, MANO and ours results are also obtained upon an I2L-MeshNet pipeline. 
    }
    \label{tab:FreiHAND}
\end{minipage}
\hfil
\begin{minipage}[h]{0.4\textwidth}
    \centering
    \includegraphics[width=1\linewidth]{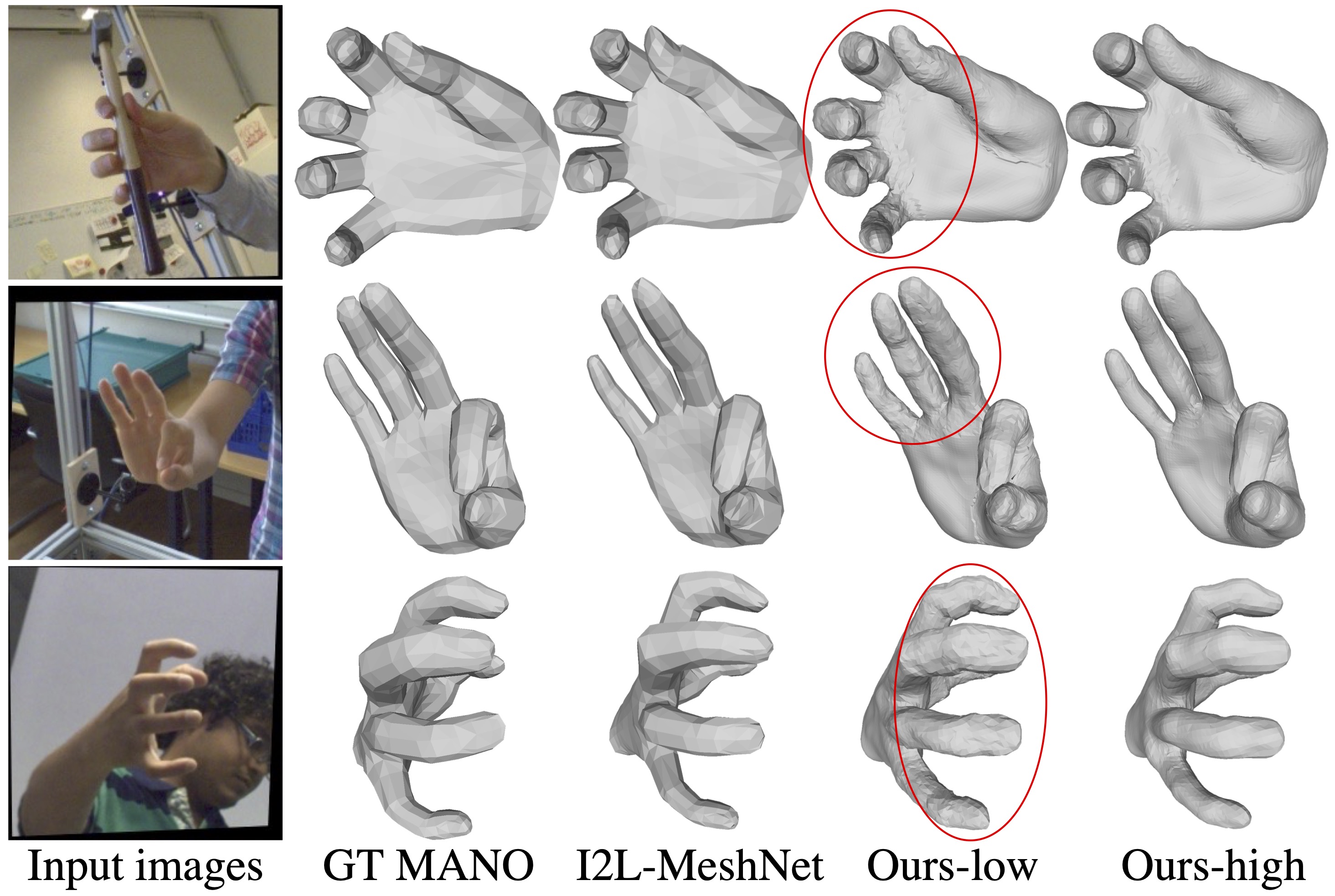}
    \vspace{-0.7cm}
    \captionof{figure}{Results of shape reconstruction from images on FreiHAND.
    }
    \label{fig:FreiHAND}
\end{minipage}
}}
\vspace{-0.2in}
\end{table*}

\subsection{Results}
We validated our method on three tasks using different datasets and compared it with state-of-the-art baselines.
\noindent \textbf{Skeleton-driven hand reconstruction.} In this task, we set NASA \cite{deng2020nasa} and HALO \cite{karunratanakul2021skeleton} as our baselines. NASA represents an articulated shape with part-based implicit functions, and HALO builds upon NASA and specializes in neural hand representation. HALO proposes to eliminate demands for GT bone transformation by using differentiable skeleton canonicalization with 3D keypoints. Following HALO, we conduct experiments on YT3D \cite{kulon2020weakly}. As shown in Table~\ref{tab:K2M-YT3D}, our method consistently outperforms HALO and NASA in all metrics. Note that we fixed the latent shape codes $\gamma$ (Section~\ref{sec:pmit}) to be fully skeleton-driven, as HALO, to ensure a fair comparison. 
As shown in Fig.~\ref{fig:YT3D}, our method achieves more accurate and smoother outputs in articulated hand modeling than HALO.
We attribute the artifacts of HALO to their rigid skeleton canonicalization, which can not handle non-rigid deformation at connecting spots of parts effectively.

\noindent \textbf{Hand reconstruction from point clouds.} 
To evaluate the accuracy of our 3D hand modeling, we test our model on high-quality point cloud data using the MANO test set \cite{romero2022embodied}. This dataset includes poses and identities that differ from the training data used to develop our model. 
In Table~\ref{tab:PF-MANO}, we compare our results with several state-of-the-art approaches, including MANO \cite{romero2022embodied}, image-learned methods (such as VolSDF \cite{yariv2021volume}, NASA \cite{deng2020nasa}, NARF \cite{noguchi2021neural}, and LISA-im \cite{corona2022lisa}) that are trained with multi-view images with the help of volume rendering, and another scan-learned method (LISA \cite{corona2022lisa}). 
Our method outperforms the MANO model, which uses the same 1,544 scans for training, as well as all image-learned methods. We also outperform LISA-full and achieve comparable results to LISA-geom \cite{corona2022lisa}. However, it is worth noting that LISA-geom and LISA-full are trained on a private dataset (3DH) consisting of 13k scans with optional multi-view images, while our method uses nearly 10 times fewer data. This demonstrates our method's superior ability to model high-fidelity hand meshes and generalize well.
We provide visualizations of some reconstructions and error heatmaps in Fig.~\ref{fig:MANO_test}. Our method produces accurate and high-fidelity hand reconstructions.

\begin{table*}[tb]
\centering
\vspace{-0.2in}
\makebox[0pt][c]{\parbox{1\textwidth}{
\begin{minipage}[h]{0.28\textwidth}
    \centering
    \setlength{\abovecaptionskip}{0.cm}
    \centering
    \scalebox{0.93}{
    \begin{tabular}{c|cc}
    \hline
    Method & $\rm P_{err}$ $\downarrow$ & $\rm M_{err}$ $\downarrow$ \\ \hline
    MANO \cite{romero2022embodied}              & 8.51           & 3.75           \\ \hline
    DHM \cite{moon2020deephandmesh}             & 3.45           & 1.98           \\ \hline
    \revise{Ours}                  & \textbf{3.14}           & \textbf{1.50}           \\ \hline
    \end{tabular}
    }
    \vspace{0.2cm}
    \caption{Results of shape reconstruction from images on the publically released part of the DHM dataset.
    }
    \label{tab:IR-DHM}
\end{minipage}
\hfil
\begin{minipage}[h]{0.33\textwidth}
\scalebox{0.85}{
\begin{tabular}{c|cc|cc}
\hline

\multirow{2}{*}{Configs} & \multicolumn{2}{c|}{Recon. to scan} & \multicolumn{2}{c}{Scan to recon.} \\ 
\cline{2-5} 
& V2V$\downarrow$ & V2S$\downarrow$ & V2V$\downarrow$ & V2S$\downarrow$ \\ \hline
 w/o $E_{norm}$             & 0.51                 & 0.40                & 0.51                 & 0.44                \\
 w/o \revise{D.S.C.}                   & 0.47                 & 0.36                & 0.46                 & 0.39                \\
 w/ $E_{{mano}^+}$            & 0.45                     & 0.34                    &  0.44                    &  0.38                   \\
Ours                            & \textbf{0.37}        & \textbf{0.27}       & \textbf{0.37}        & \textbf{0.31}       \\ \hline
\end{tabular}
}
\vspace{-0.1in}
\caption{Result of ablation study on MANO test set.
}
\label{tab:ablation}
\end{minipage}
\hfil
\begin{minipage}[h]{0.35\textwidth}
\vspace{-0.1in}
\centering
\includegraphics[width=0.9\linewidth]{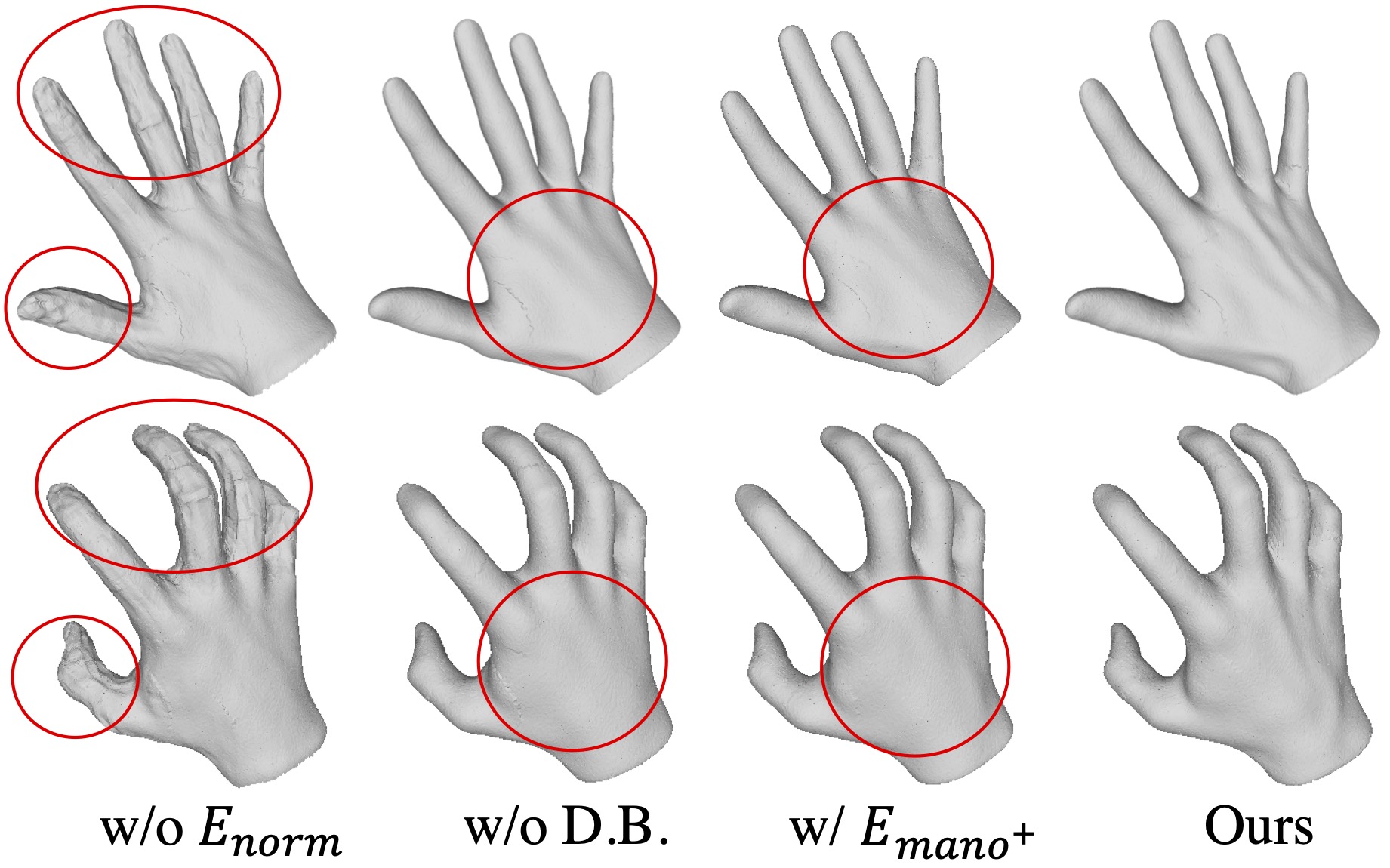}
\vspace{-0.1in}
\captionof{figure}{Qualitative results of ablation study. 
}
\label{fig:ablation}
\end{minipage}

}}
\vspace{-0.1in}
\end{table*}

\noindent \textbf{Single-view hand reconstruction.} 
Many existing methods for reconstructing 3D hands from single-view images only evaluate their performance on benchmarks with coarse MANO mesh annotations, which fail to capture the fine-detailed 3D reconstructions achievable with more advanced techniques such as DHM \cite{moon2020deephandmesh} and our approach.
To address this limitation, we evaluate our method on the DHM dataset, which provides 3D reconstructions from depth observations for evaluating meticulous 3D shape predictions. 
Using the publicly available data and official implementation of multi-view training, we compare our model's single-view  reconstruction performance to DHM and its variant with our hand mesh model in Table~\ref{tab:IR-DHM}.
\revise{The result of DHM is different from that of \cite{moon2020deephandmesh} since all the methods in the Table~\ref{tab:IR-DHM} are trained and tested on the publicly available portion of DHM, as the entire dataset is not accessible.} 
Our results show that our model significantly improves the accuracy of both shape and pose predictions, demonstrating its superior ability to represent articulated hand shapes.
Moreover, our model's clear-cut shapes can also aid in accurate pose estimation, as observed by \cite{kong2022identity}. Qualitative results are presented in Fig.~\ref{fig:DHM}, confirming the efficacy of our approach.

We also evaluate the proposed method on FreiHAND \cite{zimmermann2019freihand} with MANO pose and mesh annotations. 
We use our high-fidelity parametric hand model, which was trained on the MANO dataset scans, as a differentiable layer in the I2L-MeshNet pipeline. 
We use the predicted hand skeleton output and a two-layer MLP to estimate our shape latent code.  
We refer to this setting as "Ours-low" as \revise{we directly replace our template from the RefNet with the MANO template taking advantage of their compatibility (Section~\ref{sec:refnet}), and} use the annotated MANO mesh to supervise the predicted shape with corresponding per-vertex constraints. 
We compare our results with two other parametric hand models, MANO \cite{romero2022embodied} and NIMBLE \cite{li2022nimble}, using the I2L-MeshNet pipeline. As shown in Table \ref{tab:FreiHAND}, Ours-low outperforms both MANO and NIMBLE by a significant margin. 
Despite using a lightweight MLP to predict parametric factors, our approach achieves comparable results to the state-of-the-art method \cite{moon2020i2l}, with significantly reduced network memory (by 78\%) and computation costs (by 95\%). 
Additionally, differentiable parametric models, such as MANO, NIMBLE, and our proposed method, can serve as a differentiable layer, enabling more related tasks, such as weakly-supervised learning \cite{kulon2020weakly,chen2021model} and explicit control of hand motion in videos \cite{yang2020seqhand}.
Visualizations of our results are shown in Fig.~\ref{fig:FreiHAND}, where I2L-MeshNet fits MANO-level resolution GT meshes, while Ours-low demonstrates the ability to obtain finer surfaces.

However, using coarse MANO meshes in our training resulted in rough and non-smooth hand reconstructions.  
To address this, we introduced looser supervision by using the chamfer distance between the GT MANO vertex coordinates and our hand reconstruction to supervise the network, rather than tight per-vertex constraints. 
This approach was applied to our high-resolution parametric template (Ours-high), resulting in higher-fidelity hand reconstructions from input single-view images (as shown in Fig.~\ref{fig:FreiHAND}). 
However, while Ours-high achieves state-of-the-art qualitative results, it performs slightly worse in terms of quantitative evaluation metrics in Table \ref{tab:FreiHAND}. We suspect that the vertex-wise correspondence of MANO annotations using linear blend skinning and linear deformation correction may not accurately reflect real-world deformations, which are non-rigid and non-linear.
Overall, our model can be easily integrated into existing methods to learn high-quality hand reconstructions from MANO-level annotated datasets. We achieved state-of-the-art qualitative results and competitive quantitative results under the MANO metric.

\subsection{Ablation Study}
\label{sec:ablation}
We conduct ablation studies on our high-fidelity hand model by exploring various configurations.

\noindent \textbf{Effect of the deformation field.} The proposed deformation field bridges the SDF of the deformed hand and the canonical hand. On this basis, we derive $E_{norm}$ in our training objectives. 
We ablate whether $E_{norm}$ is used during training (Ours vs. w/o $E_{norm}$ in Fig.~\ref{fig:ablation} and Table \ref{tab:ablation}), and the results show that $E_{norm}$ is crucial to guarantee a smooth and immersive hand shape and brings key improvements to the quantitative results of point cloud 3D hand reconstruction on MANO test set.

\noindent \textbf{Effect of deformation \revise{skip connections}.}
We report the results of the holistic architecture in Fig.~\ref{fig:misc} in terms of whether the deformation \revise{skip connections} are used (Ours vs. w/o \revise{D.S.C.}). As presented in Fig.~\ref{fig:ablation}, the proposed \revise{skip-connection} design more effectively represents the details of the hand surface and achieves better overall reconstruction performance (see Table \ref{tab:ablation}).

\noindent \textbf{Effect of dense correspondence supervision.} 
To compare the impact of explicit dense correspondence supervision versus implicit deformation field, we replace $E_{mano}$ in our training objectives (Section~\ref{sec:ldf}) with $E_{mano^{+}}$, which provides dense correspondence through barycentric coordinates by upsampling MANO mesh annotations by a factor of 10. 
As shown in Fig.~\ref{fig:ablation} and Table \ref{tab:ablation}, using $E_{mano^{+}}$ resulted in degraded performance, demonstrating the superiority of implicitly building dense correspondence over pseudo supervision.

\subsection{Discussion}

\begin{table}[tb]
\centering
\makebox[0pt][c]{\parbox{0.48\textwidth}{
\begin{minipage}[t]{0.48\textwidth}
    \centering
    \vspace{-0.1in}
    \includegraphics[width=\linewidth]{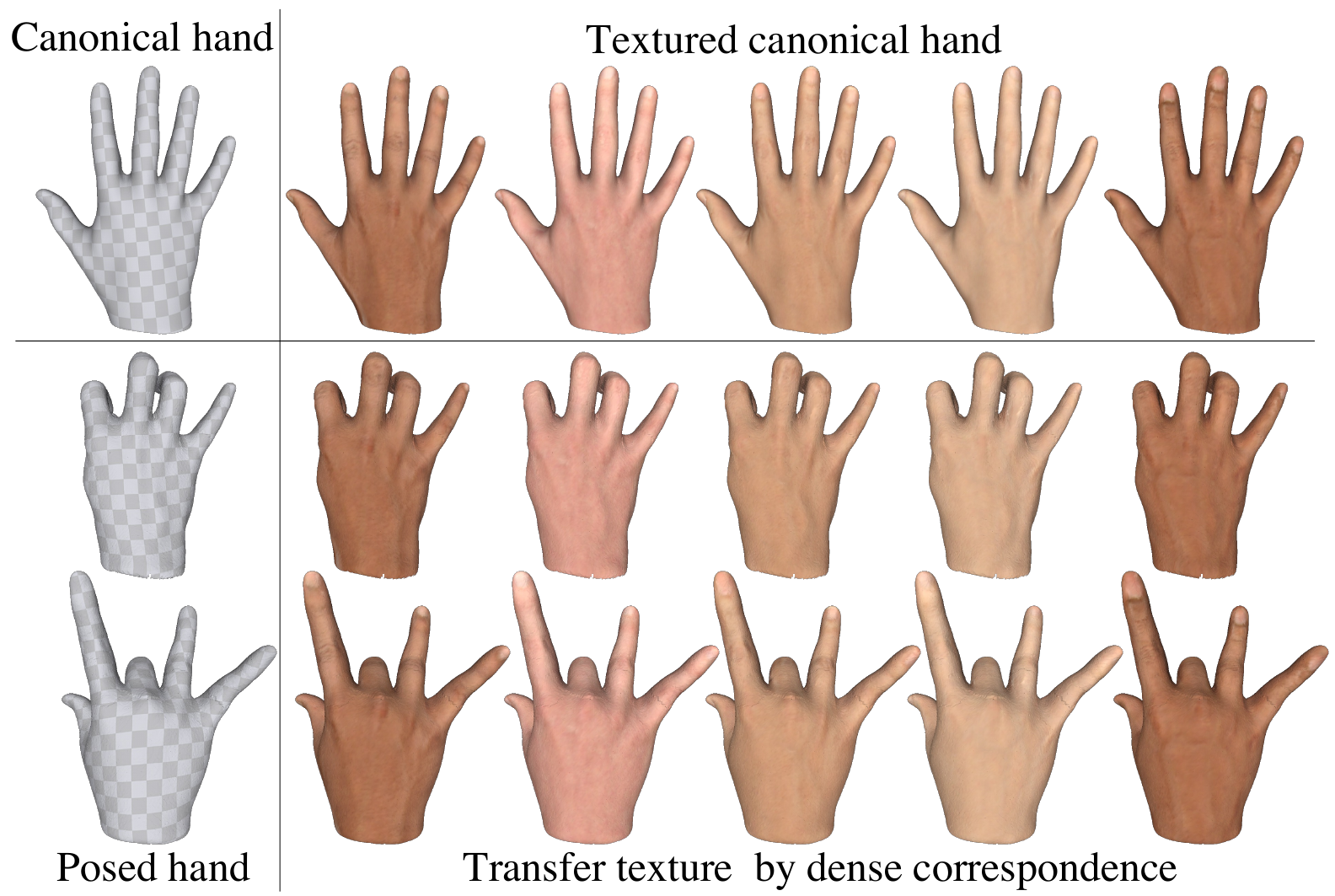}
    \vspace{-0.25in}
    \captionof{figure}{Results of transferring arbitrary hand textures through dense correspondence. We utilize HTML \cite{qian2020html} to obtain textured canonical hands.}
    \label{fig:Texture}
\end{minipage}%
\vspace{0.2cm}
\par
\begin{minipage}[t]{0.48\textwidth}
    \centering
    \scalebox{0.75}{
    \begin{tabular}{c|c|c|c|c|c}
    \hline
    Method       & MANO \cite{romero2022embodied} & DHM \cite{moon2020deephandmesh} & LISA  \cite{corona2022lisa}& HALO \cite{karunratanakul2021skeleton}  & Ours \\ \hline
    Speed (s) & 0.003  & 0.005   & $>$5     &   16.3       &  0.099    \\ \hline
    \end{tabular}
    }
    \vspace{-0.3cm}
    \caption{Comparison of inference speed. The time consumed for a feed-forward hand reconstruction is reported.
    }
    \vspace{-0.1in}
    \label{table:time}
\end{minipage}%
}}
\vspace{-0.1cm}
\end{table}

\noindent \textbf{Dense correspondence for texture modeling.} 
Our model deforms a canonical hand mesh shared by all poses and identities to obtain dense correspondences among the hand reconstructions.
We explore an application of per-vertex correspondences in transferring arbitrary textures from the texture model, such as HTML~\cite{qian2020html}, to our hand reconstructions. 
Fig.~\ref{fig:Texture} shows that our model can seamlessly combine with the HTML texture model to produce vividly textured hands with accurate geometry.

\noindent \textbf{Inference speed.} 
We compare the proposed method's inference speed with some baselines in Table~\ref{table:time}. 
\revise{We implement all methods on the same consumer device: NVIDIA RTX2080Ti GPU, Intel Xeon E5-2620 v4@2.10GHz, 11 GB main memory.}
Our method achieved significantly faster inference speed than other implicit models like LISA \cite{corona2022lisa} and HALO \cite{karunratanakul2021skeleton}, by avoiding the time-consuming Marching Cubes process. 
However, our part-based design, which improves accuracy and generalization, resulted in slightly slower inference times than other parametric mesh methods like MANO \cite{romero2022embodied} and DHM \cite{moon2020deephandmesh}. 
We believe that improving the efficiency of our part-based design can be a future research direction.

\noindent \textbf{Limitation.}
While our method achieves state-of-the-art performances in multiple tasks, it currently lacks pose priors, real-time inference capabilities, and support for queries beyond geometry modeling, such as intersection detection. 
Future research directions could include parameterizing hand skeletons, refining the part-based design to reduce computation cost and latency, \revise{modeling temporal smoothness and consistency} and extending the deformation field to enable potential one-to-many correspondence.

\section{Conclusion}
We introduce \OURS{}, a parametric hand model featuring an implicit template that generates high-fidelity hand reconstructions for various poses and identities. 
Our model offers efficient inference and dense correspondence inherited from traditional parametric meshes, as well as infinite-resolution reconstructions derived from implicit representations. 
By utilizing the proposed deformation field, we establish a dense correspondence between canonical and deformed spaces. 
Our model is fully differentiable with respect to both the skeleton and shape latent code, making it easy to incorporate into existing methods for versatile applications.
Our experiments demonstrate that \OURS{} outperforms previous methods in multiple downstream tasks.

\vspace{0.1in}
{\small
\noindent \textbf{Acknowledgements}: This work was supported by the National Natural Science Foundation of China (NSFC) under Grant 62106177, and the Natural Science Fund
for Distinguished Young Scholars of Hubei Province under Grant 2022CFA075. The numerical calculation was supported by the super-computing system in the Super-computing Center of Wuhan University.
}
{\small
\bibliographystyle{ieee_fullname}
\bibliography{iccv2023}
}

\end{document}